# Scalable Gaussian Processes for Data-Driven Design using Big Data with Categorical Factors

Liwei Wang[a,b], Suraj Yerramilli[c], Akshay Iyer[b],
Daniel Apley[c], Ping Zhu[a], Wei Chen[b]

a. The State Key Laboratory of Mechanical System and Vibration,
Shanghai Key Laboratory of Digital Manufacture for Thin-Walled Structures,
School of Mechanical Engineering,
Shanghai Jiao Tong University
Shanghai, P.R. China

b. Dept. of Mechanical Engineering
Northwestern University
Evanston, IL, USA

c. Dept. of Industrial Engineering and Management Sciences
Northwestern University
Evanston, IL, USA

**ABSTRACT**

Scientific and engineering problems often require the use of artificial intelligence to aid understanding and the search for promising designs. While Gaussian processes (GP) stand out as easy-to-use and interpretable learners, they have difficulties in accommodating big datasets, categorical inputs, and multiple responses, which has become a common challenge for a growing number of data-driven design applications. In this paper, we propose a GP model that utilizes latent variables and functions obtained through variational inference to address the aforementioned challenges simultaneously. The method is built upon the latent variable Gaussian process (LVGP) model where categorical factors are mapped into a continuous latent space to enable GP modeling of mixed-variable datasets.

By extending variational inference to LVGP models, the large training dataset is replaced by a small set of inducing points to address the scalability issue. Output response vectors are represented by a linear combination of independent latent functions, forming a flexible kernel structure to handle multiple responses that might have distinct behaviors. Comparative studies demonstrate that the proposed method scales well for large datasets with over $10^4$ data points, while outperforming state-of-the-art machine learning methods without requiring much hyperparameter tuning. In addition, an interpretable latent space is obtained to draw insights into the effect of categorical factors, such as those associated with "building blocks" of architectures and element choices in metamaterial and materials design. Our approach is demonstrated for machine learning of ternary oxide materials and topology optimization of a multiscale compliant mechanism with aperiodic microstructures and multiple materials.

1. **INTRODUCTION**

Spurred by the growth in computation capability and data resources, artificial intelligence is increasingly becoming an indispensable tool to expedite a design process and facilitate knowledge discovery in scientific and engineering problems [1]. As a non-parametric modeling approach in artificial intelligence, Gaussian processes (GPs) have come to prevail in the arena of surrogate modeling with a wide range of applications in engineering designs, such as emulating responses of expensive simulations [2], model calibration [3], sensitivity analysis and uncertainty quantification [4]. However, Gaussian processes have limitations when applied to complex design problems with challenging characteristics, such as large datasets, categorical design variables, and multiple responses. Multiscale metamaterial systems design is one such example. It requires numerous on-the-fly homogenization calculations for each new metamaterial system design due to a large number of unit cells (microstructures) considered and the nested iterations in multiscale design. In this case, data-driven design methods can greatly accelerate the design process by using an inexpensive machine learning model to replace the costly on-the-fly homogenization [5]. However, the design of such systems often involves categorical variables, such as the type of microstructure configurations and the choice of constituent materials [6, 7], that span an enormous combinatorial design space which can easily lead to exponential growth in the size of the database. Meanwhile, homogenized properties of interest for these metamaterials, e.g., stiffness tensors and thermal expansion coefficients, are examples of multi-response problems with different physical implications and behaviors for each response, lacking obvious distance metrics to describe their discrepancies. There is a need to extend GP modeling to address the obstacles caused by big data, categorical inputs, and multiple responses.

Considerable progress has been made in the literature to address each of these three challenges separately. To accommodate big data, various scalable GPs have been proposed [8]. Depending on the nature of the approximations, they can be broadly classified into two types. Globally approximated models focus on constructing an approximated covariance matrix with lower complexity and storage requirement by selecting a subset of the training data [9, 10], discarding uncorrelated entries to form a sparse covariance matrix [11], or employing some reduced-rank structures for the covariance matrix [12]. In contrast, locally approximated models deploy the divide-and-conquer strategy by considering only a subset of training data in the neighborhood of the query point to compute the predictions [13]. Meanwhile, to handle categorical factors and multiple responses, various modified covariance structures have been proposed. For example, different categories are viewed as different responses with simplified covariance structures [14, 15] while non-separable covariance structures are devised to describe multiple responses [16-19].

Recently, attempts have been made to simultaneously accommodate big data and multiple outputs [20]. However, it is not straightforward to extend these methods to handle categorical factors since the existing frameworks for categorical inputs are usually incompatible with those for big data or multiple outputs. For example, locally approximated GPs require a distance metric defined in the input variable space to obtain a subset around the query point. However, defining appropriate distance metrics for categorical input spaces is challenging. Therefore, to the best of the authors' knowledge, no existing method can simultaneously address all three challenges.

In this study, we propose a scalable latent variable GP (LVGP) modeling approach that can simultaneously accommodate a large data set, categorical factors, and multiple outputs. Specifically, as shown in Figure 1, the proposed model integrates three GP variants to handle each of the challenges, respectively, under one unified latent-variable framework [21].

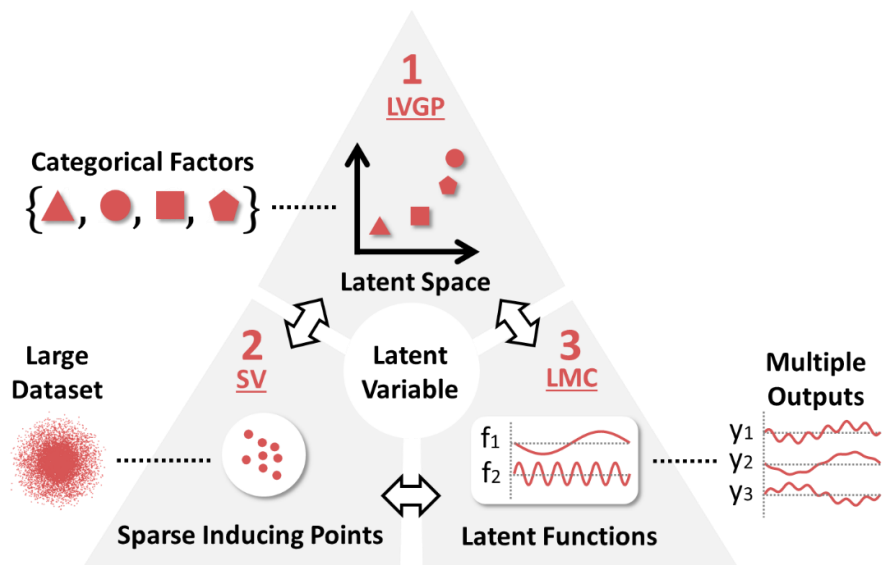

**FIGURE 1**: Three aspects integrated into the proposed Gaussian process model.

First, we adopt our previously proposed LVGP model to handle categorical variables [5, 22, 23] by mapping them into a continuous latent space to capture their joint effects on the responses. Second, to address the challenge of big data, a sparse variational (SV) approach is employed to replace the large dataset with sparse underlying inducing points to significantly reduce computation and storage complexity [24]. Finally, we model multiple outputs using a combination of independent latent functions, which is known as the linear model of coregionalization (LMC) [18].

The above three GP variants are combined into one unified GP modeling framework for large datasets with categorical inputs and multiple responses. While large data GP modeling for multi-response problems has been achieved using variational LMC [20], our contribution lies in extending the sparse variational concept to LVGP by defining inducing points in the latent space. Additionally, we propose two latent space structures for extending the LMC model to LVGP. The new synthesized GP model from this work has the following desirable features:

- **Generalizability**: Conventional correlation functions for GP modeling of continuous quantitative inputs can be readily applied to the dataset with categorical factors by using the latent variable representation. The model is also flexible for accommodating multiple responses.
- **Scalability**: The model can easily handle a large dataset with $n = 10^4 \sim 10^5$ data points in our case studies, reducing the complexity from $O(n^3)$ to $O(n_I^3)$ with the number of induing points $n_I \ll n$.
- **Accuracy**: We demonstrate in our study that the proposed model outperforms some of the state-of-the-art machine learning models, such as neural networks and boosted trees [25].
- **Interpretability**: A highly interpretable latent space of categorical variables obtained from the proposed approach provides substantial insights into the black-box problem.

This synthesized GP model is useful for a wide range of data-driven engineering design applications that involve a combinatorial design space with mixed variables and multiple responses. The aforementioned multiscale metamaterial system design is such an example of complex engineering designs, which will be demonstrated in our case studies. Other possible applications include the discovery of new molecules with different combinations of atoms and the design of composite components with various choices of architectures and constituents that result in a combinational search over mixed (categorical and quantitative) variables.

It should be noted that physics-informed machine learning (PIML) [26] is emerging as another promising tool to improve the generalizability and interpretability of models, and it has been successfully applied to many design applications [27-31]. However, fundamental differences exist between our method and these physics-informed models, in terms of application scope and functionality. PIML mainly focuses on solving PDE functions and requires prior knowledge or some reduced-order physical models [32]. In contrast, our model targets general design cases for which prior knowledge and efficient reduced-order models are unavailable. Also, our method is intended to address the relation between categorical inputs and multiple responses based on larger datasets. In contrast, PIML methods mainly consider PDE-related systems with quantitative inputs and a small or even no training dataset.

The remaining paper is organized as follows. In Section 2, we provide a brief overview of the conventional Gaussian process modeling and explain its limitations with large datasets, categorical factors and multiple outputs. Three aspects of the proposed approach are described in Section 3 by presenting three corresponding GP variants. Integration of these variants in developing a synthesized GP model is presented in Section 4. In Section 5, to validate the effectiveness, we compare the proposed method with some state-of-the-art machine learning models on two numerical examples, and two engineering examples: one on multi-response machine learning for ternary oxide materials, and another on the data-driven design of aperiodic metamaterial systems. We conclude in Section 6 and discuss the scope for future applications.

## 2. REVIEW OF GAUSSIAN PROCESS MODELING

In this section, we provide an overview of GP modeling and explain the challenges posed by mixed variables, large datasets, and multiple responses. For a single-output computer simulation model $y(\pmb{x})$ with only quantitative inputs $\pmb{x} = \{x_1, x_2, \ldots, x_p\} \in R^p$, we assume $y(\pmb{x})$ is a realization of a stochastic process:

$$Y(\pmb{x}) = \pmb{h}^T(\pmb{x})\pmb{\beta} + G(\pmb{x}), \qquad (1)$$

where $\pmb{h}(\pmb{x})$ is the prior mean function comprised of a vector of pre-defined basis functions $\pmb{h}(\pmb{x}) = [h_1(\pmb{x}), \ldots, h_m(\pmb{x})]^T$, $\pmb{\beta} = [\beta_1, \ldots, \beta_m]^T$ is a vector of unknown weights for basis functions and $G(\pmb{x})$ is a stationary multivariate Gaussian process with its covariance function defined as

$$cov\big(G(\pmb{x}), G(\pmb{x}')\big) = \sigma^2 r(\pmb{x}, \pmb{x}'), \qquad (2)$$

where $\sigma^2$ is the prior variance and $r(\cdot,\cdot)$ is the correlation function. Among numerous existing correlation functions, the Gaussian correlation function is commonly used:

$$r(\boldsymbol{x},\boldsymbol{x}') = \exp\{-(\boldsymbol{x}-\boldsymbol{x}')^T \boldsymbol{\Phi}(\boldsymbol{x}-\boldsymbol{x}')\}, \tag{3}$$

where $\boldsymbol{\Phi} = diag(\boldsymbol{\phi})$ and $\boldsymbol{\phi} = [\phi_1, \phi_2, \ldots, \phi_p]^T$ are scaling parameters to characterize the variability of the sample functions. The construction of a GP model requires estimating the hyper-parameters $\boldsymbol{\beta}$, $\boldsymbol{\phi}$ and $\sigma^2$ based on the size-n training dataset with input $\mathbf{X} = \{\boldsymbol{x}^{(1)}, \boldsymbol{x}^{(2)}, \ldots, \boldsymbol{x}^{(n)}\}^T$ and output $\boldsymbol{y} = \{y^{(1)}, y^{(2)}, \ldots, y^{(n)}\}^T$. A common way to determine the GP model parameters is to find a point estimate via maximum likelihood estimation (MLE). Herein, we assume a constant prior mean function with $\boldsymbol{h}^T(\boldsymbol{x})\boldsymbol{\beta} = \beta$ for the GP model. The corresponding log-likelihood can be given after ignoring the constants:

$$L_{ln}(\boldsymbol{\phi}, \beta, \sigma^2) = -\frac{1}{2}\ln|\boldsymbol{K}(\boldsymbol{\phi})| \\ -\frac{1}{2\sigma^2}(\boldsymbol{y}-\boldsymbol{1}\beta)^T \cdot \boldsymbol{K}(\boldsymbol{\phi})^{-1} \cdot (\boldsymbol{y}-\boldsymbol{1}\beta), \tag{4}$$

where $\ln(\cdot)$ is the natural logarithm, $\boldsymbol{1}$ is an $n \times 1$ vector of ones, and $\boldsymbol{K}$ is the $n \times n$ covariance matrix with $K_{ij} = \sigma^2 r(\boldsymbol{x}^{(i)}, \boldsymbol{x}^{(j)})$ for $i, j = 1, \ldots, n$. The hyperparameters are estimated by maximizing (4). With these estimated hyperparameters $\hat{\sigma}^2, \hat{\beta}$, and $\widehat{\boldsymbol{\phi}}$, the prediction $\hat{y}(\boldsymbol{x}^*)$ at any $\boldsymbol{x}^*$ can be obtained as:

$$\hat{y}(\boldsymbol{x}^*) = \hat{\beta} + \boldsymbol{r}^T \boldsymbol{K}^{-1}(\boldsymbol{y}-\boldsymbol{1}\hat{\beta}), \tag{5}$$

where $\boldsymbol{r}(\boldsymbol{x}^*) = [r(\boldsymbol{x}^*, \boldsymbol{x}^{(1)}), r(\boldsymbol{x}^*, \boldsymbol{x}^{(2)}), \ldots, r(\boldsymbol{x}^*, \boldsymbol{x}^{(n)})]^T$. The posterior covariance between the responses at the two given data points $\boldsymbol{x}^*$ and $\boldsymbol{x}'$ is obtained as:

$$cov(y^*, y') = \hat{\sigma}^2 r(\boldsymbol{x}^*, \boldsymbol{x}') - \boldsymbol{r}(\boldsymbol{x}^*)^T \boldsymbol{K}^{-1} \boldsymbol{r}(\boldsymbol{x}'), \tag{6}$$

For more detailed illustrations and implementation of the GP modeling, readers are referred to [33].

As discussed in Section 1, this conventional Gaussian process will encounter various obstacles when applied to a large dataset with categorical inputs and multiple outputs. Firstly, existing correlation functions are devised for quantitative variables and fail to accommodate categorical variables. For example, the correlation function in Equation (3) relies on a distance metric defined for input variables to describe the correlation between responses at different data points. However, discrete categories of categorical inputs only serve as a nomenclature without any well-defined distance metric. Secondly, GP models suffer from prohibitive computational costs and storage requirements on large datasets, due to computing $\boldsymbol{K}^{-1}$ and $|\boldsymbol{K}|$ in Equation (4). The subsequent computational and storage complexities are $O(n^3)$ and $O(n^2)$, respectively. Thirdly, it is not trivial to extend this GP model for multiple outputs obtained from simulators that jointly simulate different types of quantities [17]. While training an independent single-response GP model for each output is straightforward, it entails a time-consuming training process, especially for large datasets. Also, if the correlation between outputs is poorly captured, the GP model will result in a poor prediction power and inappropriate joint uncertainty representation.

## 3. VARIANTS OF GAUSSIAN PROCESSES FOR ADDRESSING DATA CHALLENGES

In this section, we discuss three GP variants to address the data challenges associated with categorical factors, big data, and multiple outputs, respectively. These variants are all built upon the concept of latent representation, including the LVGP model for handling categorical inputs, a sparse variational GP (SVGP) model with inducing points for managing big data challenges, and a GP model with the linear model of coregionalization (LMC) to predict multiple outputs. These three variants will be integrated to form the proposed scalable LVGP approach in the next section.

### 3.1. LVGP Model for Categorical Factors

Different categories of categorical variables lack a well-defined distance metric, which precludes the use of conventional kernels devised for quantitative variables. However, as illustrated in mapping A of Figure 2, for a physical model, there are always some underlying quantitative physical variables that explain the effects of any categorical factor on the response(s). Space spanned by these (perhaps extremely high-dimensional) underlying physical variables induces a natural distance metric between different categories of the categorical variables. Therefore, according to sufficient dimension reduction arguments [34, 35], we could assume a low-dimensional latent space to capture the joint effects of these underlying variables, as shown in mapping B of Figure 2. Based on this insight, we recently proposed an LVGP model to enable GP modeling for a dataset with categorical inputs [5, 23]. This method has been shown to have advantages over state-of-the-art counterparts, such as GPs with unrestrictive covariance [36], multiplicative covariance [15], and additive covariance [14], in terms of predictive power and model interpretability [23].

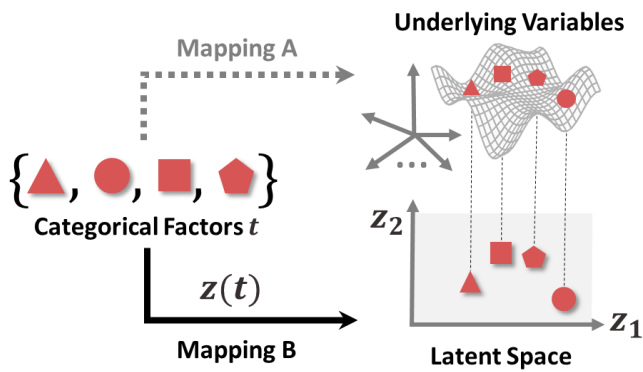

**FIGURE 2**: Illustration of the latent variable representation for categorical factors. The shape here represents the categorical factors for geometry design.

Specifically, consider a single-response computer simulation model $y(u)$ with input $u = [x^T, t^T]^T$ containing both quantitative variables $x = [x_1, x_2, ..., x_p]^T \in R^p$ and categorical variables $t = [t_1, t_2, ..., t_q]^T$, with the $j^{th}$ categorical factor $t_j \in \{1, 2 ..., l_j\}$, where $l_j \in N^+$ is the total number of categories for $t_j$. By assuming a $g$-dimensional latent vector $z_j(t_j) = [z_{j,1}(t_j), ..., z_{j,g}(t_j)]^T \in R^g$ for each $t_j$, the original mixed-variable input $x$ can be transformed into quantitative input vector $s = [x^T, z(t)^T]^T \in R^{p+q*g}$, where $z(t) = \left[z_1(t_1)^T, \ldots, z_q(t_q)^T\right]^T$. The standard GP model can then be modified as (using a constant mean function):

$$Y(s) = \beta + G(s), \tag{7}$$

$$cov(G(s), G(s')) = \sigma^2 r(s, s'), \tag{8}$$

Since the transformed input vector $s$ contains only quantitative variables, we can use any existing correlation function in equation (8). Herein, we still adopt the prevailing Gaussian correlation function:

$$r(s, s') = \exp\{-(x - x')^T \Phi (x - x') - (z - z')^T \Phi_z (z - z')\}. \tag{9}$$

It should be noted that this correlation function contains two sets of parameters to be estimated: scaling parameters $\Phi$ for quantitative variables and the set of latent vectors mapped from the categorical variables $Z = \cup_{i=1}^{q} \{z_i(1), \ldots, z_i(l_i)\}$. The scaling parameter matrix $\Phi_z$ for latent variables is fixed to be an identity matrix in LVGP since these scaling factors are absorbed into the estimated latent variable values $Z$. In our previous work [23], we follow the same procedure in Section 2 to estimate the values of $\beta$, $\sigma^2$, $\Phi$, and $Z$ via MLE.

LVGP enables easy integration with Bayesian optimization, which has been successfully applied in materials discovery and design [22, 37]. However, like conventional GPs, LVGPs also require enormous computation and storage resources when applied to big data. Moreover, the original LVGP could only accommodate a single response instead of multiple responses. To address these, we need to integrate LVGP with the two GP variants introduced in the following subsections.

### 3.2. SVGP for Big Data

In this subsection, we introduce the concept of the sparse variational (SV) model where an *artificial* training dataset that is much smaller than the original training set is used to provide approximately equivalent covariance information. These *artificial* training points, also called *inducing points*, might not be observed in the original training data and are not necessarily obtained from a real physical model. Instead, the locations and responses of these inducing points are estimated by stochastic variational inference [21] from the collected big data. This type of model, which is also called the sparse variational model [24], was demonstrated in [8] to have a good balance between scalability and predictive power across a variety of examples.

Consider a large training dataset with quantitative input data $\mathbf{X} = [\mathbf{x}^{(1)}, \mathbf{x}^{(2)}, \ldots, \mathbf{x}^{(n)}]^T$ and observed response data $\mathbf{y} = [y^{(1)}, y^{(2)}, \ldots, y^{(n)}]^T$, where $n$ is the size of the training data. In constructing the conventional GP model in Section 2, we assume a unified multivariate Gaussian distribution for the residual process $\boldsymbol{G}(\cdot)$ at $n$ training input data points $\mathbf{X}$ and $n_*$ query input data points $\mathbf{X}^*$:

$$\begin{bmatrix} \boldsymbol{G}(\mathbf{X}^*) \\ \boldsymbol{G}(\mathbf{X}) \end{bmatrix} = \begin{bmatrix} \boldsymbol{G}_* \\ \boldsymbol{G} \end{bmatrix} \sim \mathcal{N}\left(\mathbf{0}, \begin{bmatrix} K_{**} & K_{*X} \\ K_{*X}^T & K_{XX} \end{bmatrix}\right), \tag{10}$$

where $K_{*X}$ is an $n_* \times n$ cross-covariance matrix between responses at $\mathbf{X}^*$ and $\mathbf{X}$, $K_{**}$ is an $n_* \times n_*$ covariance matrix for $\mathbf{X}^*$, and $K_{XX}$ is an $n \times n$ covariance matrix for $\mathbf{X}$. $K_{XX}$ plays an essential role in both the training and prediction stages, as shown in Equations (4) through (6). We are using the covariance information of the training data stored in $K_{XX}$ to predict responses at $\mathbf{S}^*$. In other words, $\boldsymbol{G}_*$ at the query points $\mathbf{X}^*$ can be "*explained*" by $\boldsymbol{G}$ at the training points $\mathbf{X}$, as illustrated in the first row of Figure 3.

However, as discussed in Section 2, the use of this $n \times n$ covariance matrix $K_{XX}$ is the primary contributor to the curse of dimensionality in GP modeling. To address this issue, we assume that there is a small set of *inducing points* at the location $\mathbf{X}_I$ ($n_I \ll n$), with the residual process $\boldsymbol{G}(\mathbf{X}_I)$ subjects to

$$\begin{bmatrix} \boldsymbol{G}(\mathbf{X}) \\ \boldsymbol{G}(\mathbf{X}_I) \end{bmatrix} = \begin{bmatrix} \boldsymbol{G} \\ \boldsymbol{G}_I \end{bmatrix} \sim \mathcal{N}\left(\mathbf{0}, \begin{bmatrix} K_{XX} & K_{XI} \\ K_{XI}^T & K_{II} \end{bmatrix}\right), \tag{11}$$

as illustrated in the second row of Figure 3. Following the same logic in Equation (10), $\boldsymbol{G}$ at the size-$n$ training dataset $\mathbf{X}$ can be "*explained*" by $\boldsymbol{G_I}$ at the size-$n_I$ inducing input data points $\mathbf{X}_I$. The inducing points can now replace the original data to improve efficiency.

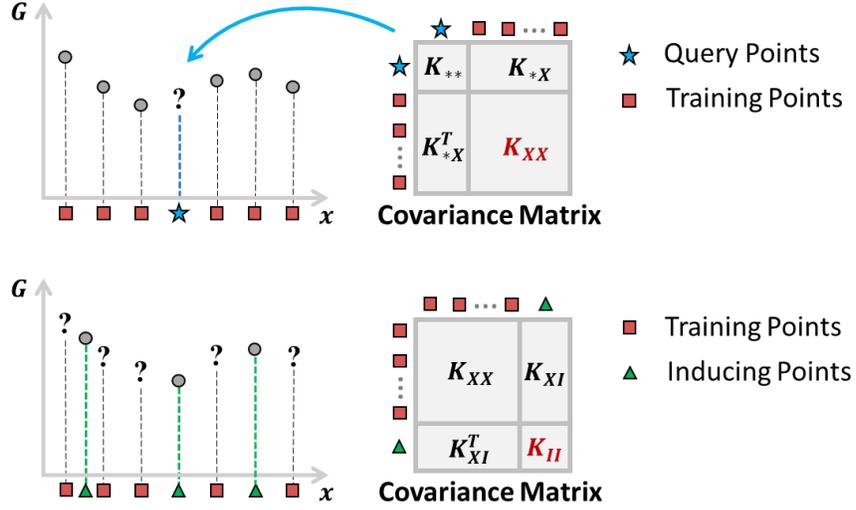

**FIGURE 3**: Covariance matrices used to describe residual process G at query points based on training points (first row) and describe G at the training points based on sparse inducing points (second row).

Under this setting, besides the original parameters in the LVGP model, we also need to estimate the locations and the corresponding $\boldsymbol{G_I}$ of these inducing points during the training process. To achieve this, a variational distribution is defined to approximate the posterior:

$$q(\boldsymbol{G}, \boldsymbol{G_I}) = p(\boldsymbol{G}|\boldsymbol{G_I})q(\boldsymbol{G_I}), \qquad (12)$$

where $q(\boldsymbol{G_I}) = \mathcal{N}(\boldsymbol{G_I}; \boldsymbol{\mu}, \boldsymbol{\Sigma})$ is the probability density function of the marginal variational distribution and $p(G|G_I)$ is the conditional distribution that is readily obtained from Equation (11). With these, parameters to be estimated include $\beta$, $\sigma^2$, $\boldsymbol{\Phi}$, $\boldsymbol{\mu}$, $\boldsymbol{\Sigma}$, $\mathbf{X}_I$ and $\boldsymbol{G_I}$. Since maximizing the likelihood function will involve the costly calculation of $\boldsymbol{K_{XX}^{-1}}$ and $|\boldsymbol{K_{XX}}|$, we turn to estimate parameters by maximizing the evidence lower bound (ELBO):

$$ELBO = L_t - D_{KL}[q(\boldsymbol{G}, \boldsymbol{G_I})||p(\boldsymbol{G}, \boldsymbol{G_I})], \tag{13}$$

with the likelihood term $L_t$ and the Kullback–Leibler (KL) divergence $D_{KL}[q(\boldsymbol{G}, \boldsymbol{G_I})||p(\boldsymbol{G}, \boldsymbol{G_I})]$ given as

$$L_t = \int log[p(\boldsymbol{y}|\boldsymbol{G})] \cdot \mathcal{N}(\boldsymbol{G}; \boldsymbol{A\mu}, \boldsymbol{A\Sigma A^T} + \boldsymbol{B})d\boldsymbol{G}, \tag{14}$$

$$D_{KL}[q(\boldsymbol{G}, \boldsymbol{G_I})||p(\boldsymbol{G}, \boldsymbol{G_I})] = \frac{1}{2}\left\{log\left(\frac{|\boldsymbol{K_{II}}|}{|\boldsymbol{\Sigma}|}\right) - n_I\right\}$$
$$+ \frac{1}{2}tr(\boldsymbol{K_{II}^{-1}\Sigma}) + (\boldsymbol{0} - \boldsymbol{\mu})^T \boldsymbol{K_{II}^{-1}}(\boldsymbol{0} - \boldsymbol{\mu}), \tag{15}$$

where $\boldsymbol{A} = \boldsymbol{K_{IX}K_{II}^{-1}}$ and $\boldsymbol{B} = \boldsymbol{K_{XX}} - \boldsymbol{K_{XI}K_{II}^{-1}K_{XI}^T}$. From Equations (13) ~ (15), we note that the evaluation of ELBO does not involve the expensive calculation of $\boldsymbol{K_{XX}^{-1}}$ and $|\boldsymbol{K_{XX}}|$. Instead, it only requires $\boldsymbol{K_{II}^{-1}}$ and $|\boldsymbol{K_{II}}|$ with the calculation complexity reduced to $O(n_I^3)$. The storage requirement can be reduced to $O(n_b^2)$ by using mini-batch stochastic gradient descent algorithms, where $n_b \ll n$ is the size of mini-batch. After the training, prediction at query points $\boldsymbol{X}^*$ can be readily obtained as

$$\hat{\boldsymbol{y}}(\boldsymbol{X}^*) = \hat{\beta} + \boldsymbol{K_{*I}K_{II}^{-1}}(\boldsymbol{\mu} - \boldsymbol{1}\hat{\beta}),$$
$$cov(\boldsymbol{X}^*, \boldsymbol{X}') = (\boldsymbol{K_{*I}K_{II}^{-1}})\boldsymbol{\Sigma}(\boldsymbol{K_{*I}K_{II}^{-1}})^T + \boldsymbol{K_{**}} - \boldsymbol{K_{*I}K_{II}^{-1}K_{*I}^T}. \tag{16}$$

The prediction in Equation (16) only depends on the sparse inducing points and thus remains efficient even with a large training data set. While this model only considers quantitative inputs, we extend it to accommodate datasets with mixed-variable inputs in Section 4.1.

### 3.3. LMC for Multi-type Responses

In this section, we introduce the linear model of coregionalization (LMC) approach to handle multiple responses [18]. The key idea behind LMC is to represent a multivariate

Gaussian process by a linear combination of independent univariate Gaussian processes. Consider a multi-response computer simulation model $y(x)$ with output $y = [y_1, y_2, \ldots, y_{N_{op}}]^T \in R^{N_{op}}$. Assume the prior model for the outputs is constructed from a linear transformation $W \in R^{N_{op} \times L}$ of $L$ ($L \leq N_{op}$) independent latent functions $f(x)$:

$$Y(x) = \beta + G(x) = Wf(x), \tag{17}$$

where $\beta$ is a vector of prior means, $G = [G_1, G_2, \ldots, G_{N_{op}}]^T$ is a multi-response stationary Gaussian process, $f(x) = \{f_l(x_l)\}_{l=1}^L$ and $f_l(x_l)$ is an independent Gaussian process with its covariance defined to be:

$$cov_l(f_l(x_l), f_l(x_l')) = \sigma^2 r_l(x_l, x_l'), \tag{18}$$

where $r_l(\cdot,\cdot)$ has the same definition as in Equation (3). By using this LMC structure, the covariance of multi-response stationary Gaussian process $G$ is given by

$$cov\left(G_i(x), G_j(x')\right) = \sum_{l=1}^{L} W_{il} cov_l(f_l(x_l), f_l(x_l')) W_{jl}, \tag{19}$$

This can be written in matrix form as

$$K_{XX'}(G(X), G(X')) = \sum_{l=1}^{L} K_{l,XX'} \otimes T_l, \tag{20}$$

where $T_l = W_{:,l} W_{:,l}^T$ with $W_{:,l}$ being the $l^{\text{th}}$ column of $W$, $\otimes$ is the Kronecker product. To estimate parameters in the LMC model, we can follow a similar approach in Section 2 to obtain MLEs (see [16] for the details). We chose this model over other alternatives, such as convolved Gaussian Processes [16, 38], due to its relative ease of training and its compatibility with the other GP variants integrated into the proposed model, which will be further illustrated in Section 4.2.

## 4. SCALABLE MULTI-RESPONSE LATENT VARIABLE GAUSSIAN PROCESS

In this section, we illustrate how the three GP variants are integrated for scalable multi-response latent variable Gaussian process modeling. We first extend the sparse variational inference to LVGP, enabling scalable modeling on a large dataset with categorical factors. This sparse variational LVGP (SV-LVGP) is then generalized to multiple responses by integrating LMC models with specially devised latent spaces of the categorical variables.

### 4.1. Extension of Variational Inference to LVGP

As mentioned in Section 3.2, the essence of the SV model is to approximate the covariance information with a set of inducing points. In the LVGP model, the original inputs $\boldsymbol{u} = [\boldsymbol{x}^T, \boldsymbol{t}^T]^T$ with both quantitative $\boldsymbol{x}^T$ and categorical factors $\boldsymbol{t}^T$ are transformed to quantitative inputs $\boldsymbol{s} = [\boldsymbol{x}^T, \boldsymbol{z}(\boldsymbol{t})^T]^T$ by mapping categories of categorical factors $\boldsymbol{t}$ to the corresponding latent vectors $\boldsymbol{z}$. As a result, there are two different input spaces $\boldsymbol{u}$ and $\boldsymbol{s}$ that can be used to define the locations of inducing points, as illustrated in Figure 4.

For the former (shown in the left column of Figure 4), the variational inference process for the inducing points will become a mixed-variable optimization problem that is computationally expensive and sensitive to initialization. Therefore, we define the locations of inducing points in the transformed quantitative input space, as shown in the right column of Figure 4. We denote the locations of inducing points, transformed training points, and query input data points as $\boldsymbol{S}_I$, $\boldsymbol{S}$ and $\boldsymbol{S}^*$, respectively. The SVGP defined in Equations (10) ~ (16) can be introduced into LVGP by simply replacing $\boldsymbol{X}_I$, $\boldsymbol{X}$ and $\boldsymbol{X}^*$ with $\boldsymbol{S}_I$, $\boldsymbol{S}$ and $\boldsymbol{S}^*$, respectively. The covariance matrices involved are calculated through Equations (8) and (9). We name this new integrated model as sparse variational latent variable GP (SV-LVGP). In SV-LVGP, parameters to be estimated include $\beta$, $\boldsymbol{\Phi}$, $\boldsymbol{Z}$, $\boldsymbol{\mu}$, $\boldsymbol{\Sigma}$, $\boldsymbol{S}_I$ and $\boldsymbol{G}_I$, which can be obtained by maximizing the ELBO as discussed in Section 3.2. Note that $\boldsymbol{Z}$ and $\boldsymbol{S}_I$ are simultaneously optimized in the training process. The feasibility of this practice is grounded in the observation that these two parameters are coupled together in the covariance matrices involving inducing points in ELBO. For better estimation of the inducing points $\boldsymbol{S}_I$, we can fix the latent vectors $\boldsymbol{Z}$ in the later stages of optimization and optimize only $\boldsymbol{S}_I$. This SV-LVGP model can now accommodate a large data set with categorical factors. It is highly scalable since the computational and storage complexity remain $O(n_I^3)$ and $O(n_b^2)$, respectively, with the number of induing points $n_I \ll n$.

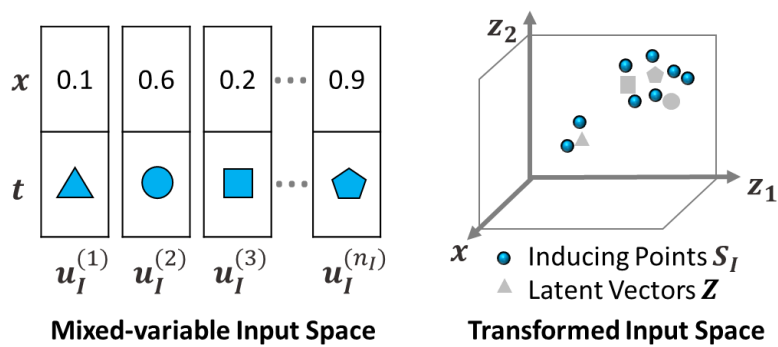

**FIGURE 4**: Defining the locations of inducing points in the mixed-variable input space (left) and the transformed quantitative input space (right).

### 4.2. Extension of LMC to SV-LVGP

In this subsection, we extend LMC to the proposed SV-LVGP model for multiple responses and present two types of model structures (illustrated in Figure 5)—one with independent latent spaces and one with shared latent spaces. Specifically, to achieve this extension, the domain of latent functions in LMC is changed from the original mixed categorical-quantitative input space to the transformed quantitative input space of SV-LVGP. In general cases, the categorical variables might show different joint effects on different responses. Accordingly, we may construct an independent latent variable space for each latent function in LMC to capture different effects of categorical variables, as shown in the first row of Figure 5.

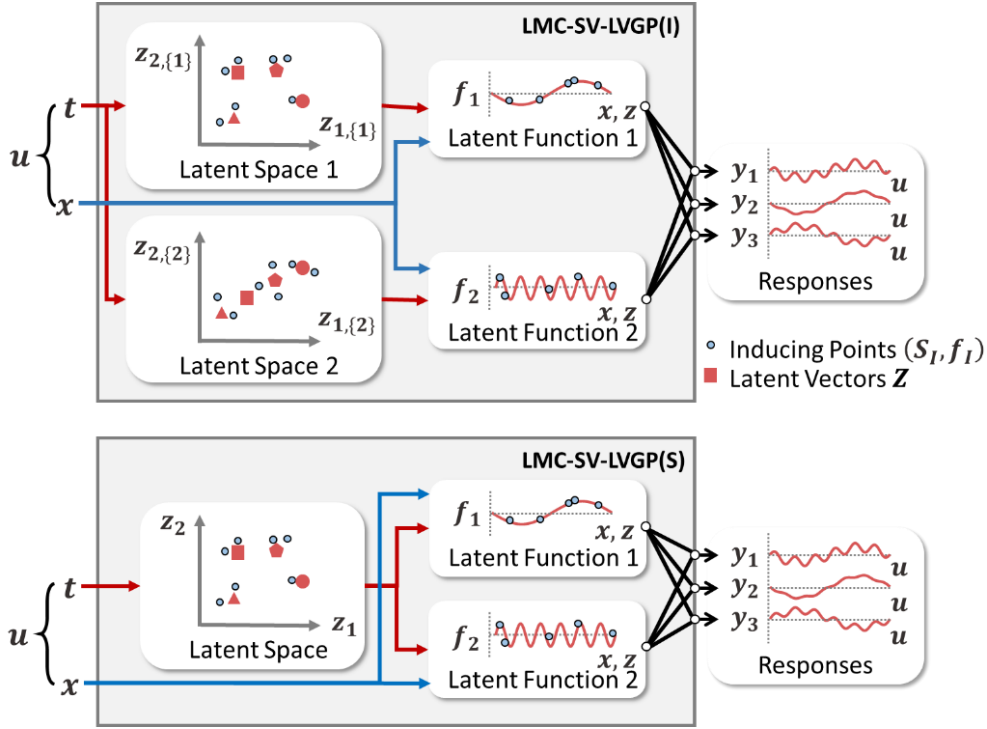

**FIGURE 5**: Illustration of the latent space structures for LMC-SV-LVGP(I) with independent latent spaces (first row) and LMC-SV-LVGP(S) with shared latent space (second row).

With this independent latent space structure, the original LMC model is changed to

$$Y(u) = \beta + G(s) = Wf(s), \tag{21}$$

where $s = [x^T, z(t)^T]^T \in R^{p+L*q*g}$ is the mapped input corresponding to $u$, $z(t) = [z_1(t_1)^T, \ldots, z_q(t_q)^T]^T \in R^{L*q*g}$ is the assembled latent vector for all categorical variables with $z_i(t_i) = [z_{i,\{1\}}(t_i)^T, \ldots, z_{i,\{L\}}(t_i)^T]^T \in R^{L*g}$, $z_{i,\{l\}}(t_i) \in R^g$ is the latent vector of $t_i$ for the $l^{\text{th}}$ latent function, $f(s) = \{f_l(s_l)\}_{l=1}^L$ with $s_l = [x^T, z_{1,\{l\}}^T, \ldots, z_{q,\{l\}}^T]^T$. The definition of the correlation function $r_l(\cdot,\cdot)$ in (18) is changed to the one for LVGP as in Equation (9). We could then follow the same procedure in Section 4.1 of introducing inducing points $(S_I, G_I)$ for scalable GP modeling, but the computational complexity will surge to $O(N_{op}^3 n_l^3)$ due to the Kronecker product in Equation (20). To avoid this significant increase in the computational costs, we propose to use the values of the latent functions $f(S_I)$ as the response data for the inducing points, instead of the final residual function $G(S_I)$. $K_{II}$ is now a block diagonal matrix, and the computational complexity is reduced to $O(Ln_I^3)$. Note that different latent functions in LMC will have different inducing points since they are defined on independent latent variable spaces. We will refer to this model as LMC-SV-LVGP(I).

In practice, one could impose constraints on the structure of the different latent variable spaces based on *prior* knowledge of the physical model to reduce the number of model parameters. For example, when the categorical variables have similar joint effects on responses, higher efficiency and interpretability can be achieved by using a special structure shown in the second row of Figure 5, in which the latent functions share the same latent variable space for all the categorical variables. Specifically, we modify the definition of the latent vector by setting $\mathbf{z}_{i,\{l\}}(t_i)^T \equiv \mathbf{z}_{i,\{1\}}(t_i)^T$ and $\mathbf{z}_i(t_i) = \left[\mathbf{z}_{i,\{1\}}(t_i)^T\right]^T \in R^g$. Moreover, we now estimate a different scaling parameter matrix $\mathbf{\Phi}_z$ for the latent variables (in Equation (9)) in different latent functions, instead of fixing them to be the identity matrix as was done earlier. These scaling parameters would account for small differences in the effects of the categorical variables on the different responses. We refer to this variant with the shared latent variable space as LMC-SV-LVGP(S). Compared to the more general model with independent latent spaces, LMC-SV-LVGP(S) sacrifices some flexibility for improving optimization efficiency with fewer parameters and inducing points to be estimated. Moreover, in the case that the categorical variables indeed have similar joint effects on different responses, the LMC-SV-LVGP(S) model will have comparable performance. We will highlight these trade-offs in the next section.

## 5. COMPARATIVE CASE STUDIES

We include two numerical examples for numerical performance comparisons and two engineering problems to demonstrate the usefulness of the proposed methods in data-driven design, including machine learning of ternary oxide materials and topology optimization of a multiscale compliant mechanism. To validate the effectiveness of our proposed methods, we compare them against two machine learning methods that are commonly used for big data: neural networks (NN) [39] and extreme gradient boosted decision trees (XGBoost) [25]. The former has been extensively used in data-driven designs due to its flexibility and capability of handling many regression problems, even with only two hidden layers [39-41]. The latter has achieved excellent results over a wide range of problems, and is recognized as a powerful tool in handling categorical and numerical inputs [42, 43], as is the case here. Consequently, these models constitute an appropriate baseline for comparison to our approach. For all case studies, 10-fold cross-validation (CV) was performed for all the models to compare their predictive power. Note that the hyperparameters of the NN and XGBoost models were tuned in an additional CV process before the comparative validation to ensure the best performance. Specifically, a random grid search with 4,000 iterations and 10-fold CV (i.e., 40,000 separate models were trained) was performed in the hyperparameter selection for NN and XGBoost, respectively, with the search space shown in Tables 1 and 2. A batch normalization layer was integrated into each hidden layer for a faster learning rate and better generalizability.

**Table 1.** The hyperparameter space of the random grid search for NN

| Number of hidden layers | Neurons per layer | Activation function (of each individual layer) | Learning rate |
| --- | --- | --- | --- |
| 1, 2, 3, 4 | 4, 8, 16, 32, 64, 128 | 'logistic', 'tanh', 'relu', 'leaky-relu', 'linear' | 0.05, 0.01, 0.005, 0.001 |

**Table 2.** The range of the random grid search for XGBoost

| Parameter | Range[*] | Parameter | Range |
| --- | --- | --- | --- |
| Colsample[**] | [0.3,0.7] | Learning rate | [0.03,0.3] |
| Gamma | (0.0,0.5] | Maximum depth | [2,6] |
| Number of estimators | [100,150] | Subsample[***] | [0.4,0.6] |

[*] uniform distribution is assumed for each range.
[**] subsample ratio of columns when constructing each tree.
[***] subsample ratio of the training instances.

In contrast, we intentionally avoid this exhaustive tuning process for all the proposed GP models to demonstrate their ease of use and generality. The proposed GP models are implemented using the GPflow package [44] in Python. The initial latent vectors for categorical variables are randomly assigned while the locations of the initial inducing points are randomly selected from the training data. We use the natural gradient optimizer [45] to optimize the variational parameters $\boldsymbol{\mu}$ and $\boldsymbol{\Sigma}$ while the Adam optimizer [46] is adopted for all other parameters for faster convergence and better parameter estimation [47, 48]. We train the GP models in batches of size 100 and set the maximum number of training iterations to 20,000.

### 5.1. Single-response Math Function

In this case study, we focus on a large single-response dataset with categorical variables generated by a math function [49] given as

$$y = \begin{cases} 7\sin(2\pi x_1 - \pi) + \sin(2\pi x_2 - \pi), & if\ t = 1 \\ 7\sin(2\pi x_1 - \pi) + 13\sin(2\pi x_2 - \pi), & if\ t = 2 \\ 7\sin(2\pi x_1 - \pi) + 1.5\sin(2\pi x_2 - \pi), & if\ t = 3 \\ 7\sin(2\pi x_1 - \pi) + 9.0\sin(2\pi x_2 - \pi), & if\ t = 4 \\ 7\sin(2\pi x_1 - \pi) + 4.5\sin(2\pi x_2 - \pi), & if\ t = 5 \end{cases} \quad (22)$$

where $x_1$, $x_2 \in [0,1]$ are continuous quantitative variables and $t \in \{1,2,3,4,5\}$ is a categorical variable with five categories representing different coefficients for the second sine function. Therefore, the true ordering of different categories should be 1-3-5-4-2 based on the second coefficient. We generate a large dataset by sampling on a $100 \times 100 \times 5$ grid in the $x_1$-$x_2$-t space, rendering 50,000 data points. To test the sensitivity of the model, we consider Gaussian random noise with three different levels of standard derivation (SD), i.e., no noise (SD=0.0), low noise (SD=0.4), and high noise (SD=4.0). We adopt a 2D latent space to represent the categorical variable in SV-LVGP, which is reported in [23] to be sufficient for most physical problems. To study the influence of the number of inducing points, we trained a set of SV-LVGP models with 50, 100, and 500 inducing points, respectively. The performance is measured by root mean squared error (RMSE), as shown in Figure 6. It should be noted that while we use normalized data during the training process with the normalized mean squared error as a loss function, the RMSE values shown in the boxplots of all the examples are mapped back to the original range of responses without normalization. Therefore, in the ideal predictive performance case, the RMSE value will equal the corresponding noise SD value with noisy data.

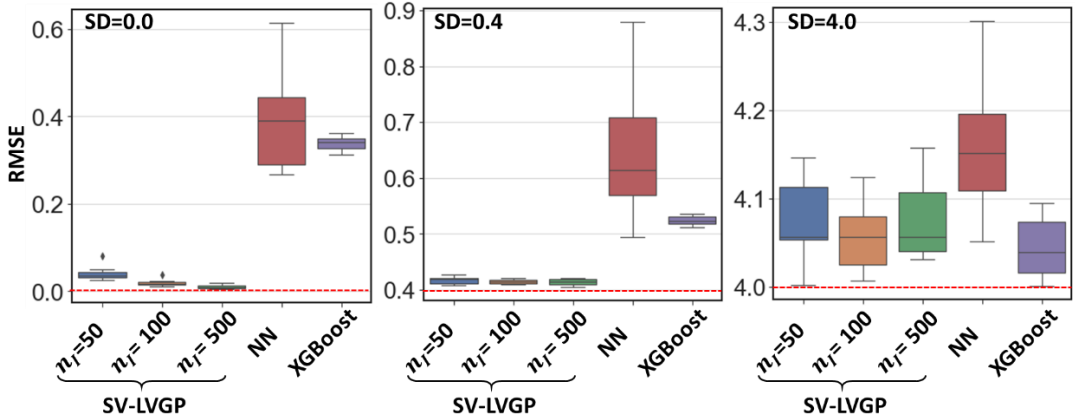

**FIGURE 6**: Boxplots of RMSE over 10-fold CV for all the models in the first case study under different noise levels. Red dashed lines represent the standard deviation of the Gaussian noise.

In no noise (SD=0.0) and low noise (SD=0.4) situations, our SV-LVGP models outperform both NN and XGBoost, even with only 50 inducing points. The reason for the poor performance of NN and XGBoost may be due to the fact that the ordering of categories does not relate to their real underlying numerical values, eliminating a critical clue for the modeling. As the number of inducing points increases, so does the predictive power of SV-LVGP. However, when the level of the noise is high (noise SD=4.0), using a larger number of inducing points does not significantly improve the prediction quality. In fact, it even results in worse performance. A possible reason is that a high level of noise in the data increases the difficulty of estimating the inducing points and therefore, models with more inducing points might require more careful initialization of these parameters and/or a more robust training procedure. Moreover, since we have intentionally skipped hyperparameter tuning for our proposed models to demonstrate their robustness to this choice, the settings of the hyperparameters we have used might not be optimal for training under high levels of noise. Although XGBoost performs the best on the highly noisy dataset, the SV-LVGP model with 100 inducing points has a similar performance. It should be noted that SV-LVGP models achieve this high accuracy without tuning their hyperparameters (which was done for NN and XGBoost), such as the learning rate and batch size. The NNs exhibit a large variance in their performances, while the SV-LVGPs have consistently better performance with much less variation across different runs. This demonstrates the robustness of the SV-LVGP model. Regarding the computational cost, the average training time is 1.2 min, 2.5 min, and 16.8 min for SV-LVGP with 50, 100, and 500 inducing points, respectively. The sparse variational model, therefore, has a manageable training expense even with over 50,000 data points. In contrast, although the NN and XGBoost models take less than a minute to train, the computational cost of the pre-tuning stage is extremely high. It took more than 18 hours to find the optimal hyperparameters in the pre-tuning stage even with parallel computing on 12 CPUs. Finally, the proposed model provides interpretation for the categories of the categorical variable through the latent space shown in Figure 7

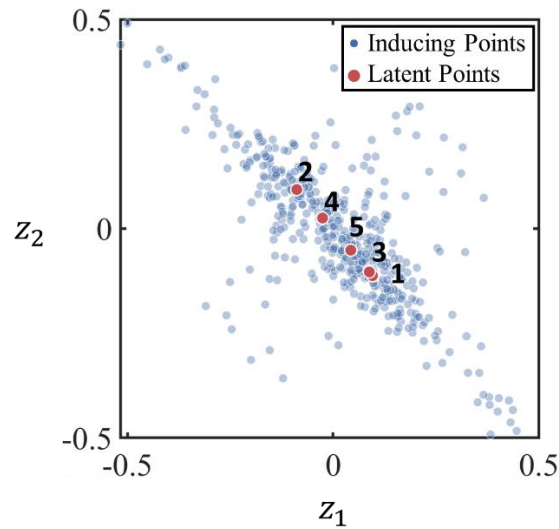

**FIGURE 7**: Latent vectors and inducing points in the latent space of SV-LVGP model with $n_I = 500$. The category of $t$ corresponding to each latent vector is marked in the figure.

. It can be seen that the latent vectors mapped from different categories reside on a straight line with a correct ordering as 1-3-5-4-2. Thus, the correlation structure captured by this mapping agrees closely with the real underlying numerical values (the coefficient of the second sine function). Therefore, even though the correlation information is lost in the categorical representation, it can be rediscovered from the data by using the proposed model. This could provide extra knowledge when applied to an unknown physical model. In contrast, NN does not have this interpretability while XGBoost fails to provide a quantitative measure for the correlation between categories. Moreover, it should be noted that the inducing points surround the latent vectors in the latent space. This is because all categorical inputs in the training data are mapped to those latent vectors. As a result, regions around the latent vectors are the most critical to describe the statistical characteristics of training data. This provides another validation of the proposed method.

### 5.2. Multi-response Math Function

In this example, we use a mathematical multi-response dataset to validate the effectiveness of the LMC-SV-LVGP model. The corresponding multi-response math function is

$$y_1 = \sum_{i=1}^{2} \frac{x_i(t_{2-i}-3)}{80} + \prod_{j=1}^{2} \cos\left(\frac{x_j}{\sqrt{j}}\right) \cos\left(\frac{50(t_j-3)}{\sqrt{2}}\right)$$
$$y_2 = \sum_{i=1}^{2} \frac{x_i(t_{2-i}-3)}{80}$$
$$+ \prod_{j=1}^{2} \cos\left(\frac{x_j}{\sqrt{j}} - \frac{(j-1)\pi}{2}\right) \cos\left(\frac{50(t_j-3)}{\sqrt{2}}\right), \quad (23)$$

where $y_1$, $y_2$ are two responses, $x_1$, $x_1 \in [-100,100]$ are continuous quantitative variables and $t_1$, $t_2 \in \{1,2,3,4,5\}$ are categorical variables with five categories. We generate a large dataset with 22,500 data points from a $30 \times 30 \times 5 \times 5$ uniform grid in the $x_1$-$x_2$-$t_1$-$t_2$ space. Similarly, we consider three levels of Gaussian noise for the dataset, i.e., SD=0.0, 0.1, 1.0. Both single-response SV-LVGP and multi-response LMC-SV-LVGP are considered in this case study. Specifically, we fit an independent SV-LVGP model for each output, which will be used as a reference for other multi-response models. For multi-response LMC-SV-LVGP, we consider three different structures: a. LMC-SV-LVGP(S) model with just a single latent function for the LMC kernel, which degenerates to the separable kernel [50], b. LMC-SV-LVGP(S) model with $L = 2$ latent functions for the LMC kernel, c. LMC-SV-LVGP(I) model with $L = 2$ latent functions for the LMC kernel. For all these models, 100 inducing points are used for the sparse variational inference. The performance of all the models over the 10-fold CV is given in Figure 8.

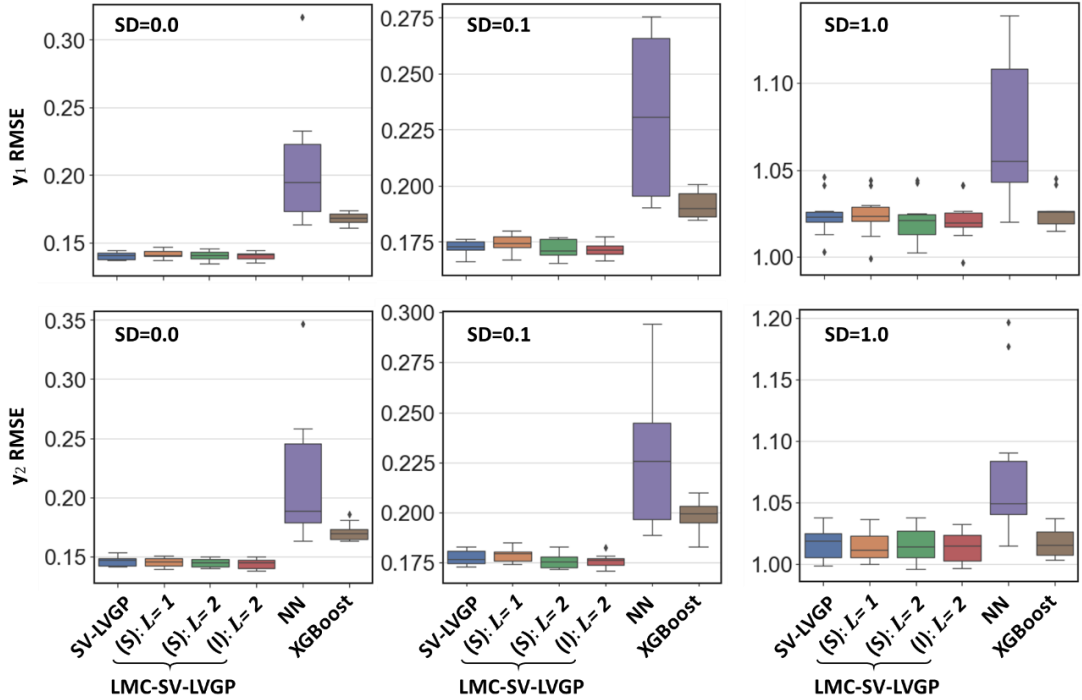

**FIGURE 8**: Boxplots of RMSE over 10-fold CV for all the models in the second case study under different noise levels. The first (second) row presents the result of the first (second) response.

It can be noted that all three LMC-SV-LVGP models have lower average RMSE values than both NN and XGBoost. The more latent functions considered in the model, the better the performance of LMC-SV-LVGP. As before, the NN shows a large variance in the predictive power across replicates, while our proposed models have a more stable performance. For this example, there is no significant difference between LMC-SV-LVGP models with shared or independent latent space, indicating the similar joint effects of categorical variables on the two responses. It is interesting to note that LMC-SV-LVGP models generally outperform the SV-LVGP model. In fact, SV-LVGP can be viewed as a special case of the LMC-SV-LVGP(I) model with the $W$ matrix restricted to be a diagonal matrix. Therefore, a more flexible structure to exploit commonalities across responses should be the reason for the better performance of LMC-SV-LVGP model over its single-response counterpart. Also, it should be noted that it takes around 8 min in total to train two separate SV-LVGP models. In contrast, it only takes 5 min and 6.5 min to train LMC-SV-LVGP models with shared and independent latent spaces, respectively. We show the latent spaces of LMC-SV-LVGP(S) and LMC-SV-LVGP(I) with $L = 2$ latent functions in Figure 9.

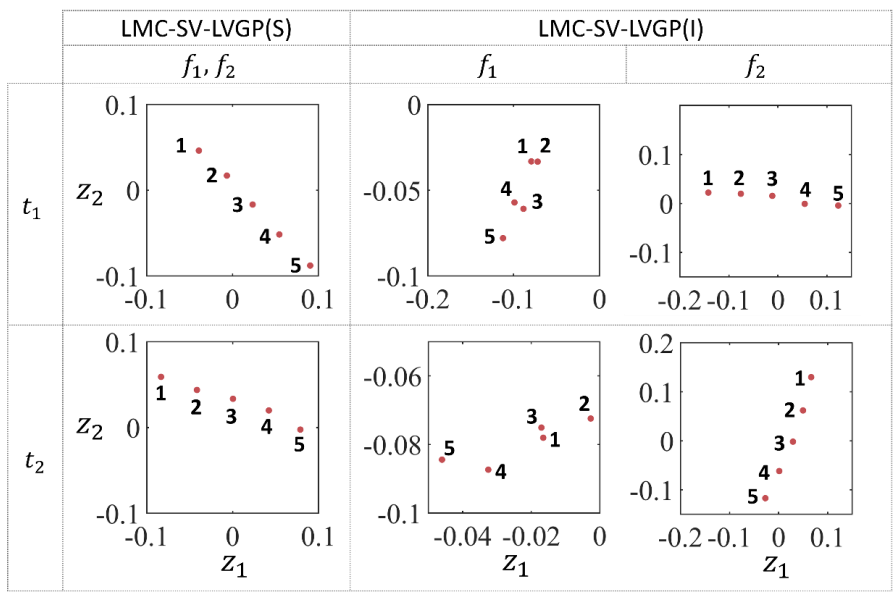

**FIGURE 9**: Latent space of LMCSV-LVGP(S) and LMC-SV-LVGP(I). The first (second) row presents the latent space for the first (second) categorical variable.

For the LMC-SV-LVGP(S) model with a shared latent space for the two latent functions, different categories of the two categorical variables are both equally distributed on a straight line and correctly ordered as 1-2-3-4-5, which again agrees with the underlying numerical $t_1, t_2$ in Equation (23). For the LMC-SV-LVGP(I) model, the two latent functions have independent latent space for the categorical variables. In this case, while similar equally spaced latent points are observed for the second latent function, the latent embedding of categorical variables for the first latent function has a very different pattern. The reason can be explained from the linear transformation for the latent functions with $W$ learned from the training process:

$$Y(u) = Wf(s) = \begin{bmatrix} -0.02 & 1.14 \\ -0.03 & 1.12 \end{bmatrix} \cdot \begin{bmatrix} f_1 \\ f_2 \end{bmatrix}, \qquad (24)$$

Note that the weights assigned to the second latent function are much larger than that of the first latent function, which indicates that the second latent function dominates the prediction result. As a result, the latent space of the second latent function captured most of the correlation information between different categories of categorical variables, implying a similar joint effect for the categorical variables on the two responses. This shows how the latent space can help to extract knowledge on the input-output relations.

### 5.3. Machine Learning for Ternary Oxide Materials

Materials informatics require a machine learning model to replace the expensive simulation or experiments in accelerating high-throughput materials discovery and iterative design process [51]. In this case study, we demonstrate that the proposed method lends itself well for use in machine learning for the combinatorial design of materials composition, by applying it to predict both formation energy and stability of ternary oxide materials. Specifically, multi-response property data for 2030 ternary oxide materials have been extracted from the Open Quantum Material Database (OQMD) [52]. These ternary oxide materials have the molecular formula as $A_{x_1} B_{x_2} O_{x_3}$, where $A$ and B can be selected from a set of 25 and 22 elements, respectively, and $O$ is the oxygen atom. $A$ and $B$ are categorical inputs, and $x_1 \sim x_3$ are quantitative inputs, forming a mixed-variable input space for the model with the formation energy and the stability as outputs. Seven models are trained on the dataset: a. SV-LVGP with 100 inducing points, b. LMC-SV-LVGP(S) model with $L = 2$ latent functions and 100 inducing points, c. LMC-SV-LVGP(I) model with $L = 2$ latent functions and 100 inducing points, d. NN, e. XGBoost, f. LMC-LVGP(I) model with $L = 2$ latent functions but no sparse variational inference, g. LMC-SV-GP model with $L = 2$ latent functions but no latent variable representation. In the LMC-LVGP(I) model, we have intentionally disabled the SV model to demonstrate its usefulness in reducing the computational expanse. It should be noted that we truncated the LMC-LVGP(I) training at 4000 iterations (which corresponded to more than 5 hours) due to excessive training time. Similarly, in the last model, we have intentionally disabled the LV component of our proposed models to show its effectiveness in handling categorical data, especially when the categorical variables have a large number of categories, as is the case here.

From their RRMSE values over 10-fold CV shown in Figure 10, it can be concluded that all three SV-LVGP models, i.e., SV-LVGP, LMC-SV-LVGP(S) and LMC-SV-LVGP(I), outperform both NN and XGBoost in predicting the formation energy and stability. Multi-response LMC-SV-LVGP models perform better than single-response SV-LVGP as before. This indicates the use of the LMC model can better accommodate multiple responses with different behaviors. Note that LMC-LVGP has a similar performance as XGBoost but much worse than that of other SV-LVGP models. Although its performance most likely would be improved if more training iterations are performed, we truncated the training at 4000 iterations ($> 5$ hours), because this is already more than two orders of magnitude larger than the training time ($< 5$ min) for all the SV-LVGP models. This demonstrates the importance and usefulness of including the SV feature. Moreover, without the latent variable representation, The LMC-SV-GP model, which does not include the LV representation, has a much worse performance than the LMC-SV-LVGP models, with its RMSE values close to that of NN. This shows that the ordering of the categories captured by the LV representation is extremely important given the larger number of categories per categorical variable.

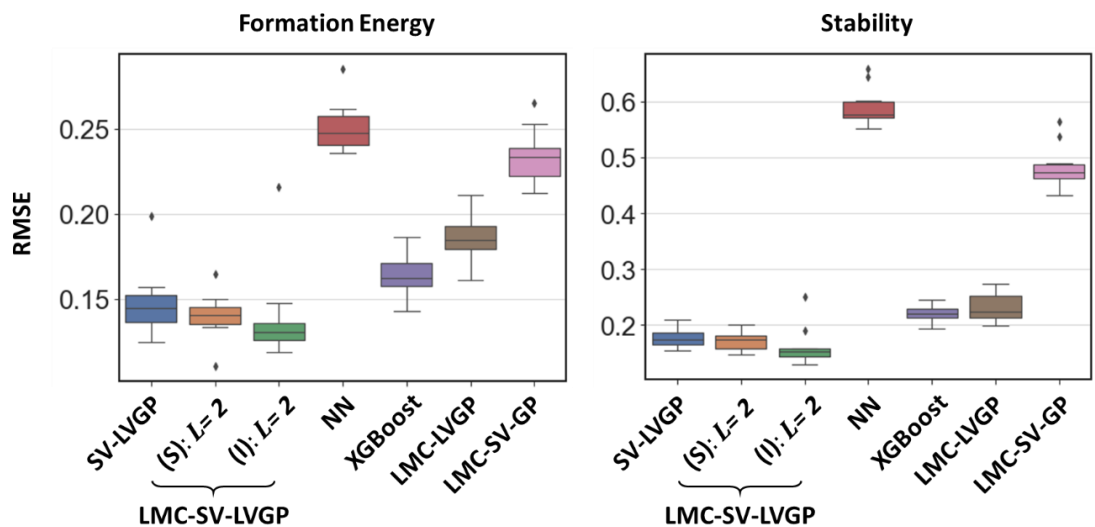

**FIGURE 10**: Boxplots of RMSE over 10-fold CV for all the models in the third case study. The left and right figures correspond to formation energy and stability, respectively.

Among the LMC-SV-LVGP models, there is a significant increase in performance when the shared space is replaced by independent latent spaces. This indicates that the type of elements included in A and B has different joint effects on the formation of energy and stability. The linear transformation for the latent functions in the LMC-SV-LVGP(I) model is

$$\begin{bmatrix} \text{formation energy} \\ \text{stability} \end{bmatrix} = \boldsymbol{W}\boldsymbol{f}(s) = \begin{bmatrix} 1.41 & 0.08 \\ 0.80 & 1.12 \end{bmatrix} \cdot \begin{bmatrix} f_1 \\ f_2 \end{bmatrix}. \tag{25}$$

The first latent function $f_1$ dominates the prediction of formation energy while the second latent function $f_2$ contributes the most to the stability prediction, indicating a large discrepancy between the two responses. We show the latent space of two categorical variables in Figure 11, which contains rich information on the effects of element types.

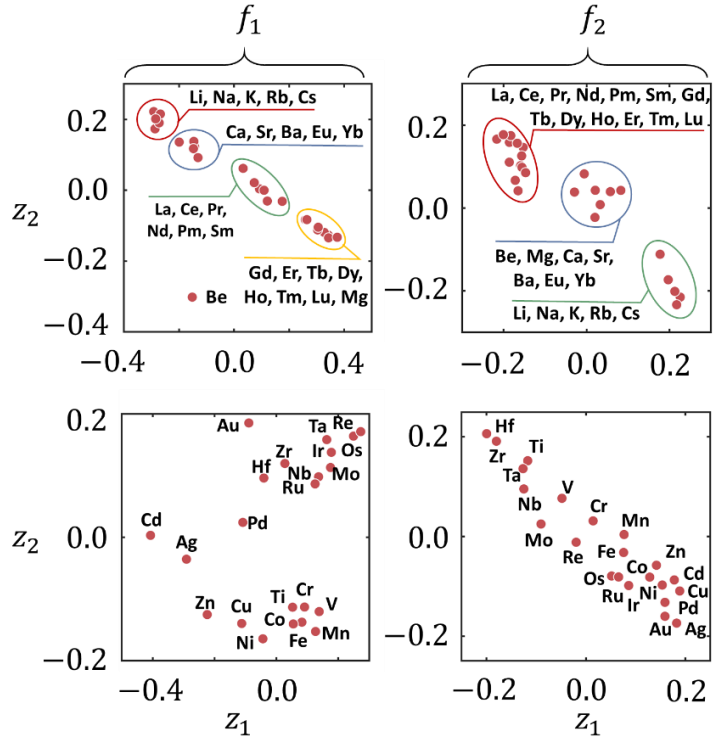

**FIGURE 11**: Latent space of LMC-SV-LVGP(I) trained on the ternary oxide materials dataset. The first (second) row shows the latent space for element A (B) in the molecular formula. The first (second) column shows the latent space used in the first (second) latent function.

For example, elements in position A form four clusters in the latent space for the first latent function. The majority of elements in each cluster belong to a specific element group in the periodic table, i.e., the alkali element group (marked by a red ellipse), the alkaline-earth element group (marked by a blue ellipse), the first and second half of the lanthanides element group (marked by a green and a yellow ellipse, respectively). Since $f_1$ dominates the prediction of formation energy, this clustering indicates that these groups have different effects on the formation energy. In contrast, there are only three clusters in the latent space for $f_2$, with the elements from the lanthanides element group being merged into the same cluster (marked by a red ellipse), indicating that all lanthanides elements have a similar influence on stability. Moreover, the proposed models require less time for the training and prediction after replacing the large data with 100 inducing points, thereby greatly reducing the time for high-throughput materials filtering or iterative design.

### 5.4. Data-driven Aperiodic Metamaterials System Design

In this case study, we demonstrate the usefulness of the proposed method in data-driven multiscale designs by applying it to a large database of unit-cell metamaterials for the design of aperiodic complex metamaterial systems [5, 53]. The microstructures are composed of two different base materials with one stiffer than the other. There are four variables to describe the microstructure of metamaterials, the volume fraction $x$ of the stiff material, the class of microstructure $t_1$, the type of stiff material $t_2$ and the type of soft material $t_3$. $x \in [0,1]$ is a quantitative input for the machine learning model while $t_1$ through $t_3$ are categorical inputs with the definition of their discrete categories shown in Figure 12. Large data is expected for such problems due to the high number of possible combinations. We generated 19,200 microstructures with precomputed stiffness tensor by uniformly sampling 100 volume fraction values $x$ for each possible combination of categorical variables. The stiffness tensor is calculated through energy-based homogenization which takes 3 hours to compute for the whole database on a single CPU (Intel i7-9750H 2.6GHz). Note that this evaluation process is only performed once for the database construction but can be applied to numerous data-driven design cases.

Independent entries of the stiffness tensor, i.e., $C_{11}, C_{12}, C_{22}$ and $C_{33}$, are viewed as outputs for the model.

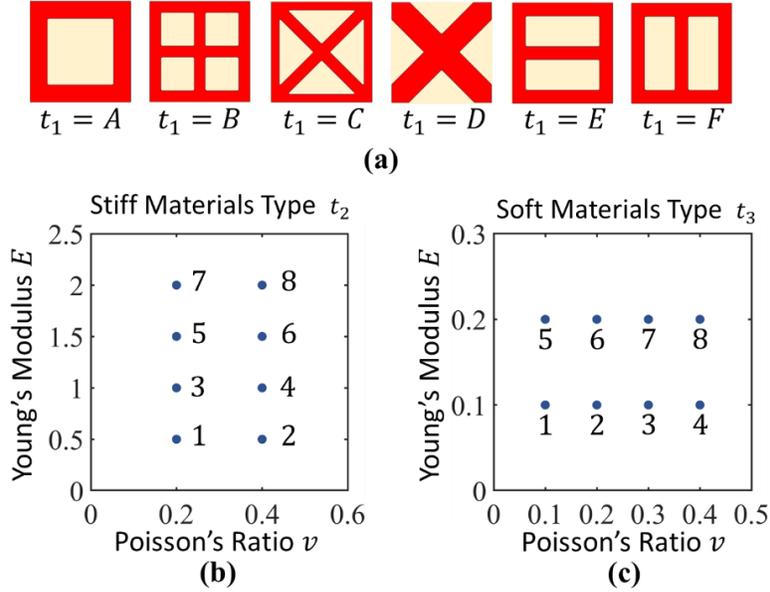

**FIGURE 12**: Categorical variables of metamaterials. (a) Microstructure classes with red and yellow regions represent the stiff and soft base materials, respectively. (b) Young's moduli and Poisson's ratios of different stiff materials. (c) Young's moduli and Poisson's ratios of different soft materials.

SV-LVGP, LMC-SV-LVGP(S), LMC-SV-LVGP(I) with four latent functions, NN and XGBoost are trained on this metamaterial dataset to compare the predictive precision, as shown in Figure 13. The three proposed models have much higher predictive power than both NN and XGBoost. While the single-response SV-LVGP model has the best performance, the difference among the three proposed GP models is not so obvious. However, as demonstrated in [5], LMC-SV-LVGP(S) is more desirable in metamaterial system design due to a much lower dimensionality of the transformed design variables (a 7D vector). Moreover, the latent space of LMC-SV-LVGP(S) provides a highly interpretable distance metric for different categorical variables, as shown in Figure 14, which will be very beneficial for the optimization process.

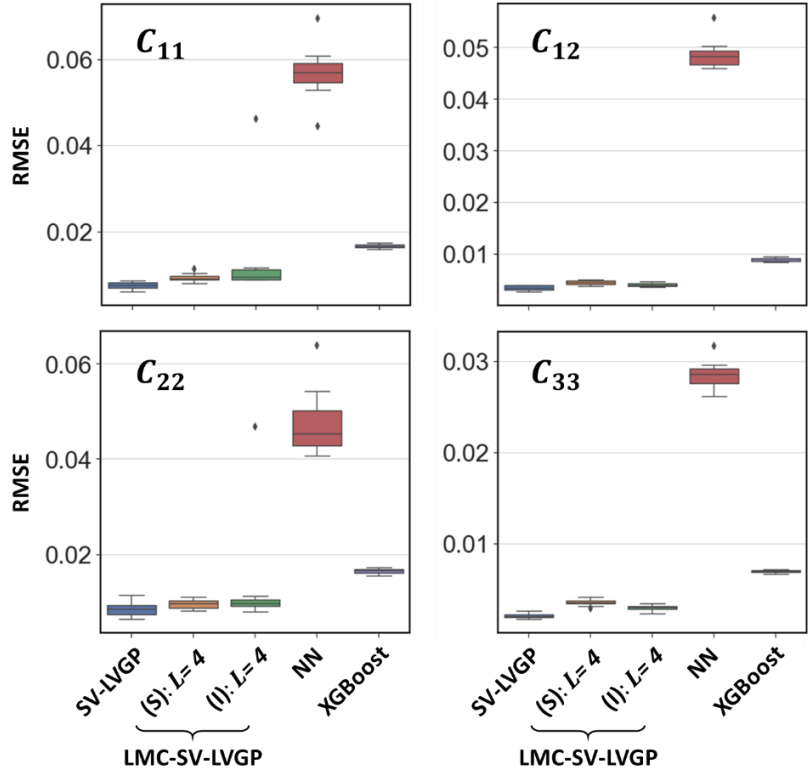

**FIGURE 13**: Boxplots of RMSE over 10-fold CV for all the models in the fourth case study. Each subfigure represents the result for an entry in the stiffness tensor.

Specifically, different classes of microstructures are distributed in a way that could reflect their similarity in the directional characteristics of the stiffness tensor. For example, classes A and B nearly overlap in the latent space shown in Figure 14 (a), which agrees with the fact that they have almost equivalent stiffness tensor under the homogenization assumption. Classes C and D are the closest neighbors for each other since they are the only pair with diagonal rods to resist shear strain. By comparing Figures 14 (b)~(c) with Figures 10(b)~(c), it is noted that the latent embeddings for the stiff and soft materials match well with the underlying values of Young's moduli and Poisson's ratios. We mark the two ascending directions for Young's modulus and Poisson's ratio in the latent space, respectively. Materials with similar Young's modulus are close to each other in the latent space. This indicates that Young's modulus has a larger impact on the stiffness tensor than Poisson's ratio.

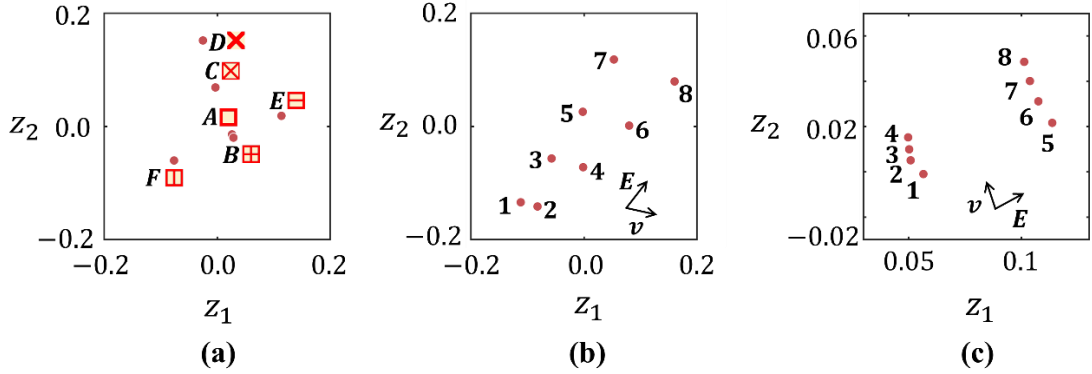

**FIGURE 14**: Latent space of LMCSV-LVGP(S) trained on the metamaterial database. (a) latent space of microstructure classes. (b) latent space of the stiff material. (c) latent space of the soft material.

To demonstrate the usefulness of the proposed method in the multiscale metamaterial systems design, we apply it in designing a multiscale compliant mechanism [54], as shown in Figure 15 (a). Consider a linear strained based actuator acting on the component, which can be modeled as a spring with stiffness $k = 0.1$ and a force $F_{in} = 1$. We aim to maximize the displacement $u_{out}$ performed on a workpiece modeled by a spring with stiffness $k$ through designing both macro- and microscale configurations. The design region is discretized into a $60 \times 40$ coarse mesh with each element filled by a microstructure discretized into a $200 \times 200$ finer mesh. The constraints imposed on the volume fraction of the stiff and soft materials are 0.3 and 0.1, respectively.

Each coarse element is associated with the aforementioned 7D transformed input vector as microscale design variables, i.e., the volume fraction $x$ of the stiff material and three sets of 2D latent vectors for the class of microstructure $t_1$, the type of stiff material $t_2$ and the type of soft material $t_3$, respectively. Each coarse element also has a macroscale topological design variable $\rho \in [0,1]$ with zero and one representing void and solid, respectively. Therefore, we only need an 8D design vector to represent the complex macro- and microscale configurations for each coarse element. In contrast, the conventional TO framework uses one-hot encoding to represent the three categorical variables, resulting in a 23D design vector for each element, i.e., one macroscale topological design variable $\rho$, 6D one-hot encoding for the class of microstructure $t_1$, and two sets of 8D one-hot encoding for the type of stiff material $t_2$ and the type of soft material $t_3$, respectively. Moreover, the dimension of the design variables will increase when more microstructure classes and materials are considered, while the design variables in our framework remain the same. This demonstrates the usefulness of the latent representation for the categorical variables in reducing the dimension of design variables.

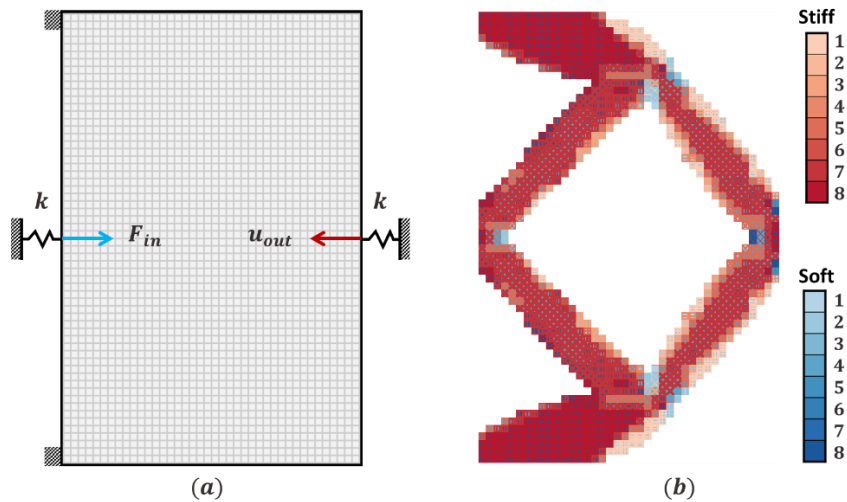

**FIGURE 15**: (a) Problem setting and (b) optimized mechanism, different types of stiff and soft materials are marked by red and blue gradient colormaps, respectively.

With the above definition, we follow the multi-scale TO framework proposed in [5] to optimize the macro-structure, the microscale configurations, and constituent materials simultaneously. Specifically, in each iteration, the proposed LMC-SV-LVGP(S) model provides the homogenized stiffness tensor and its gradient with respect to microscale design variables for each coarse element. The method of moving asymptotes [55] is then adopted to iteratively optimize the design variables based on the sensitivity value. After the optimization, the optimized multiscale design is obtained with $u_{out} = 1.3639$, as shown in Figure 15 (b). In contrast, the periodic design obtained by using the same microscale design variables for all coarse elements generates a much smaller output displacement $u_{out} = 0.8147$, highlighting the advantages of aperiodic design. Note that all eight classes of microstructures are used in the optimized structure, aligning in a way that matches with the main load-bearing directions of the macrostructure. The joint regions of different macroscale rods are composed of very soft materials, serving as hinges for the mechanism. This demonstrates the effectiveness of the simultaneous exploration of microscale configurations as well as constituent materials. Moreover, due to the low-dimensional latent variables and inexpensive LVGP model, the overall design process only takes 253 iterations and less than two minutes to converge even with 96 million fine elements in the FEA model. In contrast, the conventional aperiodic multi-scale TO needs more iterations to converge and requires around 22 minutes on the same computer platform for the on-the-fly homogenization process alone in each optimization iteration. This demonstrates that the use of the proposed machine learning model greatly accelerates the multiscale design process featuring a large combinatorial design space.

## 6. CONCLUSIONS

In this work, we have proposed a novel GP modeling approach that can accommodate big data with categorical factors and multiple responses, addressing the emerging need in AI-assisted design. The proposed model integrates three GP variants based on the concept of latent variables, which has been highlighted in this work as a powerful approach to reduce computation complexity while increasing generality and interpretability. To address the big data challenge for problems with categorical factors, we have first proposed the SV-LVGP model, which extends sparse variational inference to the LVGP for scalable mixed-variable GP modeling using inducing points. The SV-LVGP model is further generalized to cases with multiple responses by integrating the linear model of coregionalization with special latent space structures. Comparative studies demonstrate that the proposed model can easily handle $10^4 \sim 10^5$ training data points and achieve a high prediction performance that can compete with, and in most of the cases exceed, that of the state-of-the-art machine learning methods such as NN and XGBoost. The proposed model is also much easier to fit compared with these latter counterparts because it does not require a significant tuning effort. Moreover, we can gain considerable insights into the joint effects of categorical variables on the responses based on the highly interpretable latent variable space. The most remarkable demonstration of this interpretability comes from the case study for ternary oxide materials, where clusters in the latent space relate to different element groups. This differentiates our method from other conventional black-box machine learning models. Through designing a compliant mechanism, we demonstrate that the design of multiscale metamaterial systems can be greatly accelerated by using the data-driven approach and the proposed LVGP model that surrogates the material law of unit-cell structures.

For future work, we address a performance issue, wherein we had observed a drop in predictive power when more than 100 inducing points are used for highly noisy data. To resolve this issue, we plan to investigate alternative parameter initialization strategies and more robust training procedures, such as multi-start optimization. We also plan to investigate the effectiveness of the proposed models for Bayesian optimization and active learning applications involving large datasets. Nevertheless, the promising results indicate that the proposed method can be a useful tool to expedite designs where categorical variables are involved in the complex physical models or the design solutions are combinatorial in nature, such as automated design and discovery in emerging material systems.

## ACKNOWLEDGMENT

Support from the National Science Foundation (NSF) (Grant No. OAC 1835782) is greatly appreciated. Mr. Liwei Wang would like to acknowledge the support from the Zhiyuan Honors Program in Shanghai Jiao Tong University for his predoctoral study at Northwestern University.